\documentclass{article}
\usepackage{arxiv}

\usepackage{url}
\usepackage{graphicx}

\usepackage[utf8]{inputenc} 
\usepackage[T1]{fontenc}    
\usepackage{hyperref}       
\usepackage{url}            
\usepackage{booktabs}       
\usepackage{amsfonts}       
\usepackage{nicefrac}       
\usepackage{microtype}      
\usepackage{lipsum}
\usepackage[numbers]{natbib}
\usepackage{multibib}
\usepackage{verbatim}
\usepackage{amsmath}
\DeclareMathOperator*{\argmax}{arg\,max}

\newcommand\newcite{\citet}
\title{Sentence Analogies:\\
Exploring Linguistic Relationships and Regularities in Sentence Embeddings}

\author{
  Xunjie Zhu  \\
  Department of Computer Science\\
  Rutgers University\\
  Piscataway, NJ 08854 \\
  \texttt{xunjie.zhu@rutgers.edu} \\
   \And
 Gerard de Melo \\
  Department of Computer Science\\
  Rutgers University\\
  Piscataway, NJ 08854 \\
  \texttt{gdm@demelo.org}
}
\renewcommand{\today}{March 8, 2020}

\renewcommand{\vec}[1]{\mathbf{#1}}

\begin{document}
\maketitle

\begin{abstract}
While important properties of word vector representations have been studied extensively, far less is known about the properties of sentence vector representations. Word vectors are often evaluated by assessing to what degree they exhibit regularities with regard to relationships of the sort considered in word analogies. In this paper, we investigate to what extent commonly used sentence vector representation spaces as well reflect certain kinds of regularities. We propose a number of schemes to induce evaluation data, based on lexical analogy data as well as semantic relationships between sentences. Our experiments consider a wide range of sentence embedding methods, including ones based on BERT-style contextual embeddings. We find that different models differ substantially in their ability to reflect such regularities. 
\end{abstract}


\section{Introduction}
Sentence embeddings are dense vectors that reflect salient semantic properties of a sentence. Similar to how commonly used word embedding methods such as word2vec \cite{mikolov2013efficient} capture semantic relationships between words, sentence embeddings are expected to capture semantic relationships between sentences. 
A number of different sentence embedding methods have been proposed (cf.\ Section \ref{sec:embedding-methods} for an overview). More recently, pretrained language models such as ELMo \cite{peters2018deep}, BERT \cite{devlin2019bert}, XLNet \cite{yang2019XLNet}, and RoBERTa \cite{liu2019roberta} have become the method of choice when encoding text. 
Thus, such models are also often invoked to represent sentences by means of individual embeddings.\footnote{Consider e.g.\ \url{https://github.com/hanxiao/bert-as-service} and the work of Reimers et al.~\cite{reimers2019sentence}.}\ 

While important properties of word vector representations have been studied extensively, far less is known about the properties of sentence vector representations. A particularly prominent aspect of word vector representations induced by methods such as word2vec is that the vector space exhibits certain kinds of regularities. Many of these are of the sort considered in word analogies.
Proportional analogies take the form $A$ is to $B$ as $C$ is to $D$, e.g., \emph{Paris} is to \emph{France} as \emph{Berlin} is to \emph{Germany}.
\newcite{TurneyLittman2005} proposed identifying such analogies using bag-of-words vector space models.
\newcite{mikolov-etal-2013-linguistic} showed that word2vec's word vector representations reflect certain kinds of word analogies surprisingly well.
The widely used word analogy task that they proposed takes the following form.
Given embeddings $\vec{v}_A$, $\vec{v}_B$, $\vec{v}_C$, $\vec{v}_D$ for  words $A$, $B$, $C$, $D$ for an analogy of the above form, the task consists in identifying the correct word $D$ given $A$, $B$, and $C$. Most commonly, this is achieved by optimizing
\begin{align}
    \argmax_{D \in V}~~~~&\mathrm{sim}(\vec{v}_D, \vec{v}_B - \vec{v}_A + \vec{v}_C)\label{eq:word-sim}\\
    \text{\emph{s.t.}~~~~}~~&D \not\in \{A,B,C\},\nonumber
\end{align}
where $\mathrm{sim}(\vec{v}_1,\vec{v}_2)$ typically denotes cosine similarity between two vectors. 
This sort of analogy task is one of the most commonly invoked means of assessing the quality of word vector induction techniques.

However, little is known about the topology of vector representation spaces for entire sentences. In this paper we fill this gap, considering models with a dedicated sentence embedding objective as well as BERT-style pretrained embedding models. 
We study whether such sentence representation spaces as well exhibit regularities with regard to certain kinds of relationships.
To this end, we devise new datasets that are similar to typical word analogy datasets \cite{mikolov-etal-2013-linguistic}. These allow us to test empirically whether existing sentence embedding models reflect analogical relationships between sentences.

\section{Background and Related Work}

\subsection{Sentence Embedding Methods}\label{sec:embedding-methods}
In order to move from word vector representations towards representations for entire sentences, a simple baseline is to simply average the word embeddings of all words in a sentence. Although this method neglects the order of words, it performs surprisingly well in many downstream tasks. \newcite{pagliardini2017unsupervised} proposed a method to learn word and n-gram embeddings such that the average of all words and n-grams in a sentence can serve as a high-quality sentence vector. \newcite{rckl2018concatenated} improved the average pooling method by concatenating different power means of word embeddings. \newcite{almarwani2019efficient} proposed the use of a Discrete Cosine Transform (DCT) to compress word vectors into sentence embeddings. The method obtains different orders of DCT coefficients based on word embeddings and concatenates them into sentence embeddings. Unlike methods that  average word or n-grams representations, DCT considers the order and structure of words in a sentence but still maintains practical efficiency compared with RNNs or Transformer models \cite{vaswani2017attention}.

Several methods have been proposed to directly learn representations of sentences. 
The Skip-Thought Vector approach \cite{kiros2015skip}, inspired by the skip-gram word2vec approach \cite{mikolov2013efficient}, attempts to learn representations that enable the prediction of neighbouring sentences. It relies on an encoder--decoder structure based on Gated Recurrent Units.
The Quick-Thought Vector approach \cite{logeswaran2018efficient} improves both the efficiency and performance of Skip-Thought Vectors by replacing the decoder with a simple classifier that selects the correct sentence from a set of candidates.
InferSent \cite{conneau2017supervised} learns sentence representations by auxiliary supervised learning on Natural Language Inference (NLI) data, outperforming prior methods on tasks that require detailed semantic understanding \cite{zhu2018exploring}. \newcite{subramanian2018learning} proposed methods to learn general purpose sentence representations via Multi-Task Learning.

More recently, contextualized word embeddings have 
drawn considerable attention in light of the formidable gains that they achieve across a wide range of NLP and IR tasks. 
The idea of exploiting pretraining by first training a deep neural network on one task and then reusing that pretrained model on a related task had already proven incredibly fruitful in computer vision. Deep convolutional networks pretrained on ImageNet are used across a wide range of different computer vision tasks, even if the latter involve very different images than those annotated in ImageNet. 
In NLP, the pioneering work on ELMo \cite{peters2018deep} showed that significant gains can be achieved across a range of NLP tasks by considering the intermediate layers of a deep BiLSTM-based language model. 
Instead of standard bidirectional language modeling as in ELMo, the BERT approach \cite{devlin2019bert} developed at Google uses a training regime considering Cloze-style masked language modeling, in which both sides of the context are simultaneously used to reconstruct an artificially masked word, along with an additional neighbour sentence prediction task.
Although BERT has achieved great success, it neglects the dependency between masked words.  The discrepancy between pre-training and fine-tuning might also have a negative impact on the performance of some downstream tasks. Seeking to overcome these limitations of the BERT model, XLNet \cite{yang2019XLNet} is an auto-regressive Transformer-XL \cite{dai2019transformer} based model using a permutation language model as the training task. 
XLNet outperforms BERT on various downstream tasks when they share the same number of model parameters and training corpus size.
RoBERTa \cite{liu2019roberta} improves the pre-training task of the original BERT model by removing the Next Sentence Prediction task and randomly generating different masks for words in a sentence.
It also improves the performance of BERT by adding more training data.
\newcite{reimers2019sentence} proposed Sentence-BERT, which utilizes Siamese and Triplet Networks to fine-tune BERT on NLI and Semantic Textual Similarity (STS) data to obtain more semantically meaningful sentence embeddings that can be compared using cosine similarity.

\subsection{Analysing Linguistic Representations}
Whereas in the field of computer vision, there has been prominent work on understanding what is happening inside popular kinds of models \cite{ZeilerFergus2014}, 
the latent representations of recent NLP models have long remained impervious and opaque, in the sense that it is not well-understood how they represent the relevant properties of language.
While recently there has been substantial research on assessing the capabilities of BERT-like architectures \cite{RogersEtAl2020BERTology}, this research for the most part does not shed sufficient light on the topological properties of the representation space.

The most well-known way to inspect the capabilities of sentence embeddings has been via what has been dubbed \emph{probing}, i.e., supervised training of models that predict specific linguistic phenomena given embeddings as input. \newcite{kiros2015skip} evaluated the quality of their embeddings by using them for supervised downstream tasks such as sentiment polarity and question type classification. \newcite{adi2016fine} attempted to gain more specific insights by predicting word occurrences, word order, and sentence lengths. \newcite{bacon2018probing} considered this approach to predict verb tense. \newcite{ettinger2018assessing} trained classifiers for semantic roles and negation detection. \newcite{dasgupta2018evaluating} studied the argument sensitivity of the InferSent model by probing with respect to an NLI classification (contradiction vs.\ entailment). \newcite{kann2019verb} investigated verb alternation acceptability classifications. 
\newcite{conneau2018you} predicted a wide range of mostly syntactic phenomena such as major syntactic constituents, the depth of the syntactic tree, grammatical number of the subject, and grammatical number of the object. For each probing task, they provide 100,000 training instances.

Probing provides important insights about whether sufficient signals needed for a given downstream task are available if one has sufficient supervision. However, training on 100,000 instances does not reveal whether these signals are genuinely present in the sentence representations, as opposed to just being learnable from the training data. \newcite{zhu2018exploring} instead proposed assessing sentence embeddings from a relational perspective. In this paper, we specifically examine analogical relationships in terms of regularities.

\section{Methodology}

Our goal is to explore to what extent different sentence embedding spaces  reflect analogical regularities of the form $A$ is to $B$ as $C$ is to $D$.
In the remainder of this paper, we shall invoke the notation $A:B :: C:D$ to refer to this sort of relationship. We will assess such relationships using the same methods as considered for word vectors. A typical choice is the method given by Eq.~\ref{eq:word-sim} (see Section \ref{sec:metric} for further discussion). 

To be able to perform our analysis, we induce two kinds of data. In Section \ref{sec:analogies-lexical}, we create sentence analogies based on lexical analogies. In Section \ref{sec:analogy-relational}, we induce sentence analogies based on predefined relationships between sentences.

\subsection{Sentence Analogies from Lexical Analogy}
\label{sec:analogies-lexical}

The majority of sentence analogy data in our paper is induced based on  lexical analogy data. Specifically, we consider Google's word analogy dataset \cite{mikolov2013efficient} and use it to induce 5 types of semantic sentence analogies and 5 types of syntactic sentence analogy categories. 

\subsubsection{Semantic Relationships}
For semantic instances, we first create general-purpose sentence templates. Then, we replace a certain word in the template with a word from Google's word analogy dataset. We consider the following categories of relationships.

\paragraph*{Common Capital Cities.}
We first consult corpora to extract sentence templates such as ``I'm not sure if they can travel to France." Then we replace the word ``France" in the template with words from the Google Analogy dataset \cite{mikolov2013efficient} to create sentence pairs as in the following example.
\begin{quote}
$W_A$: Havana\\
$W_B$: Cuba \\
$S_A$: I'm not sure if they can travel to Havana.\\
$S_B$: I'm not sure if they can travel to Cuba.\\
\end{quote}

\paragraph*{All Capital Cities.}
We create sentence templates such as ``I've never been to Thimphu." For each word analogy pair in the Google dataset, we replace the word ``Thimphu" with pertinent words from the pairs to obtain sentence pairs as follows.
\begin{quote}
$W_A$: Amman\\
$W_B$: Jordan\\
$S_A$:  I've never been to Amman.\\
$S_B$:  I've never been to Jordan.\\
\end{quote}

\paragraph*{Currencies.}
For each currency--country pair in the Google dataset, we create different templates for currency and country, respectively. Then, we replace the target word in the currency and country templates with word pairs to generate sentence pairs as in the following example.

\begin{quote}
$W_A$: Japan\\
$W_B$: yen\\
$S_A$:  The economy in Japan was and always will be great.\\
$S_B$:  The Japanese yen appreciated due to the strong economic performance of the country.\\
\end{quote}

\paragraph*{City in State.}
We create a unified template for both city and state. Sentence pairs are then created by replacing a target word in the template with the a city or state name from the Google dataset.
\begin{quote}
$W_A$: Chandler\\
$W_B$: Arizona\\
$S_A$:  They are going down to Chandler when they get cold.\\
$S_B$:  They are going down to Arizona when they get cold.\\
\end{quote}

\paragraph*{Man -- Woman.}
We again create templates, but invoke them in more intricate ways.
For example, given a template ``My grandpa makes wooden crafts and arts.", we can replace the word ``grandpa" with any word describing family members such as ``grandma'', ``father'', and ``mother''. However, when the candidate word is a word that describes an occupation, we replace the word ``My" in the original template with "The". When the candidate word is a pronoun such as "he" or "she", we omit the word "My" from the template.
\begin{quote}
$W_A$: man\\
$W_B$: woman \\
$S_A$:  The man makes wooden crafts and arts. \\
$S_B$:  The woman makes wooden crafts and arts. \\
\end{quote}

\subsubsection{Syntactic Instances}

For syntactic questions, we first perform part-of-speech tagging and dependency parsing\footnote{We rely on SpaCy's English models for these two tasks.} to analyze the structure of the sentences in the MNLI dataset \cite{williams2018broad} and extract sentences that correspond to a certain structure. Subsequently, we invoke a set of  rules to generate new sentences from the original ones. The specific sentence generation schemes invoked to generate the evaluation data for the syntactic categories are as follows.

\paragraph*{Comparative.}
We first find a sentence containing a comparative adjective followed by ``than'', and then replace the comparative adjective with its original form and remove the noun or clause after ``than'' to obtain comparative sentence pairs of the following sort.

\begin{quote}
$W_A$: long\\
$W_B$: longer\\
$S_A$:  The second article was long.\\
$S_B$:  The second article was longer than the first article.\\
\end{quote}

\paragraph*{Nationality Adjectives.}
We create templates for nationalities and their corresponding adjectives. We then replace a target word in the template with the nationality designation or with the adjective word.

\begin{quote}
$W_A$: Egypt\\
$W_B$: Egyptian\\
$S_A$:  The man from Egypt tapped his cheek.\\
$S_B$:  The Egyptian man tapped his cheek.\\
\end{quote}

\paragraph*{Opposites.}
We first find a sentence containing an adjective and then replace the adjective with its opposite word to obtain sentence pairs of the following form.
\begin{quote}
$W_A$: possible\\
$W_B$: impossible\\
$S_A$:  It's possible to measure it.\\
$S_B$:  It's impossible to measure it.\\
\end{quote}

\paragraph*{Plurals.}
We first find a sentence with a plural noun and a numeral word between the noun, and then replace the plural noun with its singular form and replace the numeral word with word ``one", ``a", or ``an".
\begin{quote}
$W_A$: city\\   
$W_B$: cities\\
$S_A$: The Harvard data examined one city on the East coast.\\
$S_B$: The Harvard data examined 6 cities on the East coast.\\
\end{quote}

\paragraph*{Verb Conjugation.}
We first find sentences containing an auxiliary verb followed by a verb and then remove the auxiliary verb and replace the verb in the sentence with its inflected form.
\begin{quote}
$W_A$: play\\
$W_B$: plays\\  
$S_A$: Duke will play better this year.\\
$S_B$: Duke plays better this year.\\
\end{quote}

\subsection{Analogy based on Relationships Between Sentences}
\label{sec:analogy-relational}

In addition to our sentence analogy data derived from word analogies, we also created new diagnostic sentence analogy data based on specific forms of relationships between sentences. 
Our relation based analogy dataset includes two different relations: entailment and negation. Sentence pairs with entailment and negation relations are extracted from NLI datasets (including SNLI \cite{snli:emnlp2015}, Multi-NLI \cite{williams2018multi} and SICK \cite{marelli2014semeval}).

\subsubsection{Relationships}

\paragraph*{Entailment}
Given two sentence pairs $S_A$, $S_B$ and $S_C$, $S_D$, an entailment analogy holds between these two sentence pairs if the respective relationships between $S_A$ and $S_B$ and between $S_C$ and $S_D$ are both \emph{entailment}.
\begin{quote}
$S_A$: The turtle is tracking the fish.\\
$S_B$: The turtle is following the fish.\\
$S_C$: A person is dicing an onion.\\ 
$S_D$: A person is cutting an onion into pieces.\\
\end{quote}

\paragraph*{Negation}
We consider sentence pair $S_A$, $S_B$ as standing in a negation relationship if one has a negated meaning compared to the other. Given two sentence pairs $S_A$, $S_B$ and $S_C$, $S_D$, a negation analogy holds between these sentence pairs if the respective relationship between $S_A$ and $S_B$ and between $S_C and S_D$ are both \emph{negation}.
\begin{quote}
    $S_A$: There is no skilled person riding a bicycle on one wheel. \\
    $S_B$: A skilled person is riding a bicycle on one wheel.
    $S_C$: There is no man frying a tortilla.
    $S_D$: A man is frying a tortilla.
\end{quote}

\subsubsection{Candidate Sets}
For a given hypothesis, there may be a multitude of valid premises that entail it. Given an analogy of the form $S_A:S_B :: S_C:S_D$, we need to restrict the scope of candidate sentences for $S_D$ so as to ensure the uniqueness of the correct answer. Instead of considering the entire corpus as a candidate sentence set, our candidate sentence set for entailment and negation consists of one true candidate and several challenging distractor candidates that are similar to the true candidate at a superficial level but modified to be semantically different.
Table \ref{table:relation-examples} provides examples for a brief overview of the resulting task. In the following, we explain the distractor generation in further detail.

\paragraph*{Not Negation}
We insert the negation marker \emph{not} after the first auxiliary verb in the original true target sentence to generate a new distractor sentence. If the sentence already contains the negation marker \emph{not}, we instead remove it. Not Negation aims to detect whether a sentence embedding model is misled by a negation of the sentence relation caused by adding the word \emph{not}.

\paragraph*{Random Deletion}
We randomly delete words in the original sentence with a probability of 20\% to generate a new sentence. If the length of the sentence is less than 5, we delete at least one word. 
However, simply deleting arbitrary words in the original sentence may not always affect the semantic relationship. For example, consider 
the hypothesis ``John ate a yummy sandwich.'' vs.\ the premise ``John ate a delicious sandwich.'' 
If we delete the word \emph{delicious} from the premise sentence, the relation between the hypothesis and the new sentence is also entailment. In order to avoid this situation, we pick words for deletion that are not adjectives, adverbs, determiners, or auxiliary verbs.

\paragraph*{Random Masking}
Following BERT \cite{devlin2019bert}, we randomly replace tokens in the original sentence with BERT's special "[MASK]" token, where the probability of a certain token being masked is 20\%. For sentence embedding methods that are not based on BERT, the "[MASK]" token is treated as an "UNK" token, which represents unknown words. The purpose of random masking task is exploring whether replacing a word with a special meaningless token will affect a model's performance in judging a semantic relation.

\paragraph*{Span Deletion}
A number of text spans are sampled, with span lengths drawn from a Poisson distribution with $\lambda = 3$. Each text span is deleted from the original sentence. Span Deletion is inspired by the text infilling operation in BART \cite{lewis2019bart}. The only difference is that BART replaces text spans with a "[MASK]" token instead of deleting them. One difference between random deletion and span deletion is that we do not pose any restrictions on the word spans to be deleted. Since a continuous text span in a sentence often represents a phrase or a sub-clause, in most cases, the generated sentence's meaning and relationship is different from the original sentence. But there may also be some exceptions to this.
Hence, we rely on manual checking to avoid such issues.

\paragraph*{Word Reordering}
We randomly choose a word in the original sentence as a pivot, and then swap the words before and after the pivot to obtain a new sentence, which is likely grammatically incorrect and not an appropriate target to be selected. We invoke this sort of word reordering to test whether an embedding model is sensitive to semantic relation changes caused by changes of the word order. Clearly, a simple averaging of word embeddings is not able to distinguish this sort of example from the true target sentence, but it is not yet known to what extent more sophisticated sentence embedding models may suffer from this issue.

\begin{table}[]
\caption{Example of candidate sets for relation-based analogy}\label{table:relation-examples}
\centering
\begin{tabular}{lll}
\toprule
                   & Entailment                                        & Negation                                 \\ \midrule
$S_A$               & The man is heaving barbells.                       & There is no deer jumping a fence.         \\ 
$S_B$               & The man is lifting barbells.                       & A deer is jumping over the fence.         \\ 
$S_C$               & A man is singing a song and playing the guitar.    & There is no boy hitting the football.     \\ 
Positive Candidate & A man is singing and playing the guitar.           & A boy is hitting the football.            \\
\midrule
Not Negation       & A man is not singing and not playing the guitar.   & A boy is not hitting the football.        \\
Random Deletion    & A man is the guitar.                               & is the football.                          \\ 
Random Masking     & A {[}MASK{]} is {[}MASK{]} and playing the guitar. & A  {[}MASK{]} is {[}MASK{]} the football. \\ 
Span Deletion      & A man is singing the guitar.                       & A boy the football.                       \\ 
Word Reordering    & and playing the guitar A man is singing.           & The football a boy is hitting.            \\ \bottomrule
\end{tabular}
\end{table}

\section{Experiments}
In a sentence analogy task, we are given two pairs of sentences sharing a relation. For example, ``He is very enamored with culture in Egypt'' : ``He is very enamored with Egyptian culture'', and ``He is very enamored with culture in Bulgaria'' : ``He is very enamored with Bulgarian culture''.
The goal is to identify the fourth sentence given the first three sentences.
The type of analogy sought in each sentence pair is not explicitly provided. 
The number of sentence pairs and question pairs in each category of our analogy dataset is given in Table \ref{table:analogy-stat},
\begin{table}[htb]
\caption{Number of sentence and question pairs in each category}
\centering
\resizebox{0.5\textwidth}{!}{\begin{tabular}{lrr}
\toprule
Category                       & Sentence Pairs & Question Pairs \\
\midrule
Common Capital City       & 138            & 9,453           \\
All Capital Cities         & 928            & 430,128         \\
City in State        & 402            & 80,601          \\
Currency               & 150            & 11,175          \\
Family               & 126           & 7,875         \\
\midrule
Comparative            & 466            & 108,345         \\
Opposite               & 513           & 131,328         \\
Nationality Adjective & 205            & 20,910          \\ 
Plural                 & 512            & 130,816         \\ 
Verb Conjugation          & 451            & 101,475         \\ 
\midrule
Entailment       & 673            & 226,128           \\
Negation            & 511            & 130,305         \\
\bottomrule
\end{tabular}}
\label{table:analogy-stat}
\end{table}

\subsection{Evaluation Metric}\label{sec:metric}
In word analogy tasks, the offset between word vectors is often used to determine relations between words. For example, in order to solve \emph{man} is to \emph{woman} as \emph{king} is to $W$, we find a word $W$ for which the corresponding vector is the closest to $\vec{v}_\mathrm{man} - \vec{v}_\mathrm{woman} + \vec{v}_\mathrm{king}$. This amounts to optimizing Eq.~\ref{eq:word-sim}.
\newcite{levy2014linguistic} studied this in more detail, referring to the aforementioned method as \emph{3CosAdd}, while introducing a multiplicative variant called \emph{3CosMul}, which often yields better empirical results. 

\newcite{linzen-2016-issues}, \newcite{Schluter_2018_Word_Analogy_Testing_Caveat}, and \newcite{nissim2019fair} highlighted the significance of excluding the other analogy words in Eq.~\ref{eq:word-sim}.
Given a word analogy problem of the form $A:B :: C:D$, the standard procedure is to disregard any $D$ that is equal to $A$, $B$, or $C$. This constraint drastically improves the performance of word embedding models on word analogy datasets such as the Google dataset \cite{mikolov2013efficient}, but may also lead to biased results.

In our experiments, we consider both 3CosAdd and 3CosMul, and evaluate these both with the additional constraint (3CosAdd, 3CosMul) and without it (3CosAdd-U, 3CosMul-U), where the suffix -U denotes an \emph{unconstrained} evaluation.

\subsection{Embedding Methods}

In our experiments, we consider a number of embedding models. 
These include simple word vector aggregation methods such as the Average of GloVe embeddings (abbreviated as \emph{GloVe}). 
The concatenation of Discrete Cosine Transform coefficients (\emph{DCT}) embeddings are generated by concatenating the first $k$ DCT coefficients. In our experiment, $k$ ranges from 0 to 6, and for space reasons, we report the best-performing result.

For sentence embeddings based on RNNs such as Skip-Thought Vectors (\emph{SkipThought}), Quick-Thought vectors (\emph{QuickThought}), and the General Purpose Sentence Encoder by \newcite{subramanian2018learning} (\emph{GenSen}), we use the hidden state of the final RNN cell as the sentence embedding. 

For InferSent, we use max-pooling over all hidden states of RNN cells to produce sentence embeddings. Facebook released two versions of the InferSent model, the earlier version (\emph{InferSentV1}) is trained based on GloVe word embeddings, while the second version (\emph{InferSentV2}) is trained using fastText word embeddings. 

The Universal Sentence Encoder \cite{cer2018universal} comes in two versions, the first one based on Deep Average Networks (\emph{USE-DAN}), the second based on Transformer networks (\emph{USE-Transformer}). 

For contextual embedding models such as BERT,
XLNet, RoBERTa \cite{liu2019roberta}, and Sentence-BERT \cite{reimers2019sentence}, we consider two popular methods to generate a sentence embedding. The first one (\emph{-CLS}) consists in using the embedding of the special "[CLS]" token in the sentence, followed by a linear transformation and a tanh activation layer. Another method (\emph{-AVG}) involves computing the element-wise sum of contextual word representations $w_1$, $w_2$, $w_3$, ..., $w_n$ at the top level of the Transformer encoder and dividing it by the square root of the sentence length. For a given model, we only show the pooling method that obtained the highest accuracy in the experimental results. 
We consider different versions of the models (\emph{-Base}, \emph{-Large}) as released by the original authors.
In our result tables, the Sentence-BERT models by \newcite{reimers2019sentence} based on BERT and RoBERTa are referred to as \emph{SBERT} and \emph{SRoBERTa}, respectively.

\subsection{Results and Analysis}
\paragraph{Sentence Analogy from Lexical Analogy}
Table \ref{table:all-result} provides the overall aggregate results for lexical analogy-based pairs, while Table \ref{table:semantic} specifically assesses the semantic analogy categories, and Table \ref{table:syntactic} assesses the syntactic analogy categories.

With an unconstrained evaluation, we observe that all considered sentence embedding models show relatively poor success rates. We conjecture that this is because $\mathrm{sim}(D, B) - \mathrm{sim}(D, A)$ tends to be fairly small in most cases, so 3CosAdd-U and 3CosMul-U often degenerate mainly to evaluating $\mathrm{sim}(D, C)$. 

With a traditional constrained evaluation, several methods show substantial regularities. Averages of GloVe vectors draw on the linguistic regularities inherent in the GloVe vectors. The Discrete Cosine Transform method outperforms other sentence embedding methods for most of the categories. 
InferSent outperforms contextual embedding methods, with XLNet-Large obtaining the weakest results across all considered models.
Despite the strength of InferSent, fine-tuning BERT on NLI datasets does not improve the performance of the model on lexical analogy based tasks. SBERT and SRoBERTa obtained a lower accuracy than BERT and RoBERTa, respectively. 

Comparing the results of Tables \ref{table:semantic} and  \ref{table:syntactic}, we find that capturing semantic analogies is more challenging than capturing syntactic analogies. Most of the sentence embedding models we tested (except XLNet) excelled at solving syntactic question pairs using the 3CosAdd metric, while few of them perform well on semantic analogy pairs. In particular, contextual embedding-based models appear capable of reflecting syntactic phenomena, but do not appear to 
yield semantic knowledge at the same level as word embedding models such as GloVe.

\begin{table}
\caption{Experimental results on all lexical analogy-based data}
\centering
\resizebox{0.75\columnwidth}{!}{\begin{tabular}{|l|l|l|l|l|}
\hline
	 & 3CosAdd-U 	& 3CosAdd  	& 3CosMul-U 	 &3CosMul  \\ \hline
GloVe 	 & 0.4092 	 & 0.8189  	 & 0.4092  	 & 0.8039 	 \\ \hline
DCT ($k$=0) 	 & 0.5193 	 & 0.8865  	 & 0.5193  	 & 0.8688 	 \\ \hline
SkipThought 	 & 0.1805 	 & 0.6251  	 & 0.1805  	 & 0.4540 	 \\ \hline
QuickThought 	 & 0.1337 	 & 0.6318  	 & 0.1337  	 & 0.5907 	 \\ \hline
InferSentV1 	 & 0.2787 	 & 0.7118  	 & 0.2787  	 & 0.6594 	 \\ \hline
InferSentV2 	 & 0.3405 	 & 0.8323  	 & 0.3405  	 & 0.5000 	 \\ \hline
GenSen 	 & 0.3756 	 & 0.5366  	 & 0.3756  	 & 0.5038 	 \\ \hline
USE-DAN 	 & 0.0316 	 & 0.4995  	 & 0.0316  	 & 0.0658 	 \\ \hline
USE-Transformer 	 & 0.0518 	 & 0.5714  	 & 0.0518  	 & 0.0818 	 \\ \hline
BERT-Base-AVG 	 & 0.1537 	 & 0.6471  	 & 0.1537  	 & 0.6415 	 \\ \hline
BERT-Large-AVG 	 & 0.2375 	 & 0.6643  	 & 0.2375  	 & 0.5805 	 \\ \hline
XLNet-Base-AVG 	 & 0.0234 	 & 0.4223  	 & 0.0234  	 & 0.4214 	 \\ \hline
XLNet-Large-AVG 	 & 0.0105 	 & 0.2397  	 & 0.0105  	 & 0.2383 	 \\ \hline
RoBERTa-Base-CLS 	 & 0.0645 	 & 0.6117  	 & 0.0645  	 & 0.6107 	 \\ \hline
RoBERTa-Large-AVG 	 & 0.0793 	 & 0.6229  	 & 0.0793  	 & 0.6221 	 \\ \hline
SBERT-Base-AVG 	 & 0.0977 	 & 0.5108  	 & 0.0977  	 & 0.1821 	 \\ \hline
SBERT-Large-AVG	 & 0.1513 	 & 0.5347  	 & 0.1513  	 & 0.3557 	 \\ \hline
SRoBERTa-Base-AVG 	 & 0.0881 	 & 0.2548  	 & 0.0881  	 & 0.1459 	 \\ \hline
SRoBERTa-Large-AVG 	 & 0.1073 	 & 0.2978  	 & 0.1073  	 & 0.2008 	 \\ \hline
\end{tabular}}
\label{table:all-result}
\end{table}

\begin{table}
\caption{Experimental results on semantic sentence analogy pairs}
\centering
\resizebox{0.75\columnwidth}{!}{
\begin{tabular}{|l|l|l|l|l|}
\hline
	 & 3CosAdd-U 	& 3CosAdd  	& 3CosMul-U 	 &3CosMul  \\ \hline
GloVe 	 & 0.4159 	 & 0.7453  	 & 0.4159  	 & 0.7244 	 \\ \hline
DCT ($k$=1) 	 & 0.4300 	 & 0.8477  	 & 0.4300  	 & 0.8019 	 \\ \hline
SkipThought 	 & 0.2024 	 & 0.4382  	 & 0.2024  	 & 0.3563 	 \\ \hline
QuickThought 	 & 0.0367 	 & 0.3974  	 & 0.0367  	 & 0.3129 	 \\ \hline
InferSentV1 	 & 0.2655 	 & 0.6159  	 & 0.2655  	 & 0.5502 	 \\ \hline
InferSentV2 	 & 0.3755 	 & 0.7967  	 & 0.3755  	 & 0.5490 	 \\ \hline
GenSen 	 & 0.2374 	 & 0.3457  	 & 0.2374  	 & 0.2617 	 \\ \hline
USE-DAN 	 & 0.0317 	 & 0.1471  	 & 0.0317  	 & 0.0598 	 \\ \hline
USE-Transformer 	 & 0.0606 	 & 0.2773  	 & 0.0606  	 & 0.0831 	 \\ \hline
BERT-Base-AVG 	 & 0.1128 	 & 0.4481  	 & 0.1128  	 & 0.4358 	 \\ \hline
BERT-Large-AVG 	 & 0.2131 	 & 0.4852  	 & 0.2131  	 & 0.4344 	 \\ \hline
XLNet-Base-AVG 	 & 0.0239 	 & 0.1521  	 & 0.0239  	 & 0.1510 	 \\ \hline
XLNet-Large-AVG	 & 0.0062 	 & 0.0327  	 & 0.0062  	 & 0.0318 	 \\ \hline
RoBERTa-Base-CLS	 & 0.0609 	 & 0.4171  	 & 0.0609  	 & 0.4162 	 \\ \hline
RoBERTa-Large-CLS 	 & 0.0267 	 & 0.4507  	 & 0.0267  	 & 0.4496 	 \\ \hline
SBERT-Base-AVG 	 & 0.0640 	 & 0.4190  	 & 0.0640  	 & 0.1473 	 \\ \hline
SBERT-Large-AVG 	 & 0.1143 	 & 0.4898  	 & 0.1143  	 & 0.3656 	 \\ \hline
SRoBERTa-Base-AVG 	 & 0.0135 	 & 0.0347  	 & 0.0135  	 & 0.0248 	 \\ \hline
SRoBERTa-Large-AVG 	 & 0.0190 	 & 0.0886  	 & 0.0190  	 & 0.0587 	 \\ \hline
\end{tabular}
}
\label{table:semantic}
\end{table}

\begin{table}
\caption{Experimental Results on syntactic sentence analogy pairs}
\centering
\resizebox{0.75\columnwidth}{!}{
\begin{tabular}{|l|l|l|l|l|}
\hline
	 & 3CosAdd-U 	& 3CosAdd  	& 3CosMul-U 	 &3CosMul  \\ \hline
GloVe 	 & 0.4019 	 & 0.8993  	 & 0.4019  	 & 0.8908 	 \\ \hline
DCT ($k$=0) 	 & 0.5276 	 & 0.9298  	 & 0.5276  	 & 0.9305 	 \\ \hline
SkipThought 	 & 0.1565 	 & 0.8295  	 & 0.1565  	 & 0.5608 	 \\ \hline
QuickThought 	 & 0.2397 	 & 0.8882  	 & 0.2397  	 & 0.8945 	 \\ \hline
InferSentV1 	 & 0.2983 	 & 0.8547  	 & 0.2983  	 & 0.8222 	 \\ \hline
InferSentV2 	 & 0.2883 	 & 0.8853  	 & 0.2883  	 & 0.4271 	 \\ \hline
GenSen 	 & 0.5815 	 & 0.8212  	 & 0.5815  	 & 0.8645 	 \\ \hline
USE-DAN 	 & 0.0315 	 & 0.8847  	 & 0.0315  	 & 0.0724 	 \\ \hline
USE-Transformer 	 & 0.0423 	 & 0.8930  	 & 0.0423  	 & 0.0804 	 \\ \hline
BERT-Base-AVG 	 & 0.1984 	 & 0.8647  	 & 0.1984  	 & 0.8665 	 \\ \hline
BERT-Large-AVG 	 & 0.2641 	 & 0.8602  	 & 0.2641  	 & 0.7402 	 \\ \hline
XLNet-Base-AVG 	 & 0.0228 	 & 0.7177  	 & 0.0228  	 & 0.7171 	 \\ \hline
XLNet-Large-AVG 	 & 0.0152 	 & 0.4660  	 & 0.0152  	 & 0.4642 	 \\ \hline
RoBERTa-Base-AVG 	 & 0.2155 	 & 0.8531  	 & 0.2155  	 & 0.8524 	 \\ \hline
RoBERTa-Large-AVG 	 & 0.1161 	 & 0.8442  	 & 0.1161  	 & 0.8433 	 \\ \hline
SBERT-Base-AVG 	 & 0.1348 	 & 0.6119  	 & 0.1348  	 & 0.2205 	 \\ \hline
SBERT-Large-AVG 	 & 0.1921 	 & 0.5842  	 & 0.1921  	 & 0.3448 	 \\ \hline
SRoBERTa-Base-AVG 	 & 0.1703 	 & 0.4970  	 & 0.1703  	 & 0.2792 	 \\ \hline
SRoBERTa-Large-AVG 	 & 0.2046 	 & 0.5281  	 & 0.2046  	 & 0.3572 	 \\ \hline
\end{tabular}
}
\label{table:syntactic}
\end{table}

\begin{table}
\caption{Experimental results on entailment sentence analogy pairs}
\centering
\begin{tabular}{|l|l|l|l|l|}
\hline
	 & 3CosAdd  	& 3CosMul \\ \hline
GloVe 	 & 0.0137 	 & 0.0040 	 \\ \hline
DCT ($k$=6) 	 & 0.6667 	 & 0.2370 	 \\ \hline
SkipThought 	 & 0.7308 	 & 0.0768 	 \\ \hline
QuickThought 	 & 0.5785 	 & 0.2140 	 \\ \hline
InferSentV1 	 & 0.2212 	 & 0.0309 	 \\ \hline
InferSentV2 	 & 0.1263 	 & 0.0087 	 \\ \hline
GenSen 	 & 0.2122 	 & 0.0023 	 \\ \hline
USE-DAN 	 & 0.2475 	 & 0.0606 	 \\ \hline
USE-Transformer 	 & 0.1560 	 & 0.0165 	 \\ \hline
BERT-Base-AVG 	 & 0.4415 	 & 0.3744 	 \\ \hline
BERT-Large-AVG 	 & 0.6896 	 & 0.4335 	 \\ \hline
XLNet-Base-CLS 	 & 0.4700 	 & 0.4635 	 \\ \hline
XLNet-Large-CLS 	 & 0.3715 	 & 0.3700 	 \\ \hline
RoBERTa-Base-CLS 	 & 0.8703 	 & 0.8704 	 \\ \hline
RoBERTa-Large-CLS 	 & 0.5781 	 & 0.5782 	 \\ \hline
SBERT-Base-AVG 	 & 0.1781 	 & 0.0363 	 \\ \hline
SBERT-Base-CLS 	 & 0.1983 	 & 0.0396 	 \\ \hline
SBERT-Large-AVG 	 & 0.2238 	 & 0.0422 	 \\ \hline
SBERT-Large-CLS 	 & 0.2528 	 & 0.0562 	 \\ \hline
SRoBERTa-Base-AVG 	 & 0.2207 	 & 0.0454 	 \\ \hline
SRoBERTa-Large-AVG 	 & 0.2708 	 & 0.0591 	 \\ \hline
\end{tabular}
\label{table:entailment-analogy}
\end{table}

\begin{table}
\caption {Experimental results on negation sentence analogy pairs}
\centering
\begin{tabular}{|l|l|l|l|l|}
\hline
	 & 3CosAdd  	& 3CosMul \\ \hline
GloVE 	 & 0.0213 	 & 0.0056 	 \\ \hline
DCT ($k$=6) 	 & 0.5921 	 & 0.2257 	 \\ \hline
SkipThought 	 & 0.5694 	 & 0.1456 	 \\ \hline
QuickThought 	 & 0.5578 	 & 0.3423 	 \\ \hline
InferSentV1 	 & 0.0289 	 & 0.0154 	 \\ \hline
InferSentV2 	 & 0.0070 	 & 0.0237 	 \\ \hline
GenSen 	 & 0.1023 	 & 0.0037 	 \\ \hline
USE-DAN 	 & 0.1434 	 & 0.1061 	 \\ \hline
USE-Transformer 	 & 0.0234 	 & 0.0510 	 \\ \hline
BERT-Base-AVG 	 & 0.2370 	 & 0.2384 	 \\ \hline
BERT-Large-AVG 	 & 0.3588 	 & 0.2939 	 \\ \hline
XLNet-Base-AVG 	 & 0.3550 	 & 0.3514 	 \\ \hline
XLNet-Large-AVG 	 & 0.3000 	 & 0.2992 	 \\ \hline
RoBERTa-Base-CLS 	 & 0.3219 	 & 0.3217 	 \\ \hline
RoBERTa-Large-AVG 	 & 0.4633 	 & 0.4628 	 \\ \hline
SBERT-Base-AVG 	 & 0.0065 	 & 0.0499 	 \\ \hline
SBERT-Base-CLS 	 & 0.0055 	 & 0.0414 	 \\ \hline
SBERT-Large-AVG 	 & 0.0059 	 & 0.0383 	 \\ \hline
SBERT-Large-CLS 	 & 0.0050 	 & 0.0203 	 \\ \hline
SRoBERTa-Base-AVG 	 & 0.0053 	 & 0.0304 	 \\ \hline
SRoBERTa-Large-AVG 	 & 0.0095 	 & 0.0282 	 \\ \hline
\end{tabular}
\label{table:negation-analogy}
\end{table}

\paragraph{Sentence Analogy from Relationships Between Sentences}
In Table \ref{table:entailment-analogy}, we provide the results on the 
entailment sentence analogies data from Section \ref{sec:analogy-relational}. Note that because a sentence always entails itself, we omit the unconstrained versions of 3CosAdd and 3CosMul.
The RoBERTa-Base model achieves the highest accuracy on the entailment analogy task, substantially outperforming even RoBERTa-Large. It seems that the RoBERTa-Large model more often confuses the real premise sentence with the not-negated form of the original sentence. This problem also is observed in models fine-tuned on NLI datasets such as SBERT, SRoBERTa, and InferSent, which suggests that supervised training on NLI does not help with regard to the entailment analogy task. The pertinent negation signals appear to be captured in ways that are not amenable to analogical vector inference.
GloVe performs very poorly on the entailment analogy task, since it is easily misled by Random Masking. The DCT model avoids this problem by concatenating the first $k$ DCT coefficients.

In Table \ref{table:negation-analogy}, we consider the negation analogy task. DCT fares better at capturing negation analogy than other models. Most of the models we tested are easily misled by Not Negation, especially for models fine-tuned on NLI data. Recall from Table \ref{table:relation-examples} that the goal is to find a sentence $S_D$ that stands in a negation relationship with $S_C$, while the Not Negation distractor as well stands in a negation relationship with $S_D$ and hence is not an appropriate choice for $D$.

\section{Conclusion}
In this paper, we proposed several new datasets to test whether existing sentence embedding models exhibit regularities with regard to sentence analogies.
Most of the sentence embedding models we tested succeeded in recognizing syntactic analogies based on lexical ones, but were not good at capturing semantic regularities by means of an analogy task. 
Moreover, the remarkable success of BERT-style contextual embeddings does not always translate into better regularities in the vector space of fixed-length sentence embeddings. 
More training data and model parameters as well do not necessarily yield
better results. 
In many cases, word vector averages or a Discrete Cosine Transform of word embeddings outperform more complex sentence embedding models.

\bibliography{references,acl2020}  

\bibliographystyle{acl_natbib}

\end{document}